\documentclass[10pt,twocolumn,letterpaper]{article}

\usepackage{iccv}
\usepackage{times}
\usepackage{epsfig}
\usepackage{graphicx}
\usepackage{amsmath}
\usepackage{amssymb}
\usepackage{caption}
\usepackage[nocompress]{cite}

 \iccvfinalcopy 

\def\iccvPaperID{1530} 

\ificcvfinal\fi
\begin{document}

\title{The Pose Knows: Video Forecasting by Generating Pose Futures}

\author{Jacob Walker, Kenneth Marino, Abhinav Gupta, and Martial Hebert\\
Carnegie Mellon University\\
5000 Forbes Avenue, Pittsburgh, PA 15213\\
{\tt\small \{jcwalker, kdmarino, abhinavg, hebert\}@cs.cmu.edu}}
\maketitle

\begin{abstract}
    Current approaches in video forecasting attempt to generate videos directly in pixel space using Generative Adversarial Networks (GANs) or Variational Autoencoders (VAEs). However, since these approaches try to model all the structure and scene dynamics at once, in unconstrained settings they often generate uninterpretable results. Our insight is to model the forecasting problem at a higher level of abstraction. Specifically, we exploit human pose detectors as a free source of supervision and break the video forecasting problem into two discrete steps. First we explicitly model the high level structure of active objects in the scene---humans---and use a VAE to model the possible future movements of humans in the pose space. We then use the future poses generated as conditional information to a GAN to predict the future frames of the video in pixel space. By using the structured space of pose as an intermediate representation, we sidestep the problems that GANs have in generating video pixels directly. We show through quantitative and qualitative evaluation that our method outperforms state-of-the-art methods for video prediction.

\end{abstract}

\section{Introduction}
\begin{figure}
\begin{tabular}{ cc }
\includegraphics[height=0.14\textwidth,width=0.21\textwidth]{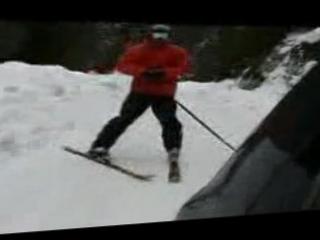} &
\includegraphics[height=0.14\textwidth,width=0.21\textwidth]{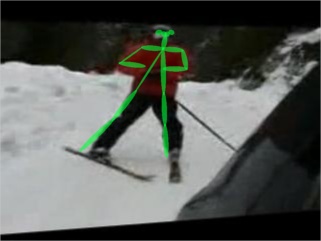} \\
{\footnotesize (a) Input Clip} &  {\footnotesize (b) Start Pose} \\
\includegraphics[height=0.14\textwidth,width=0.21\textwidth]{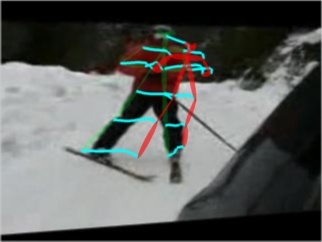} &
\includegraphics[height=0.14\textwidth,width=0.21\textwidth]{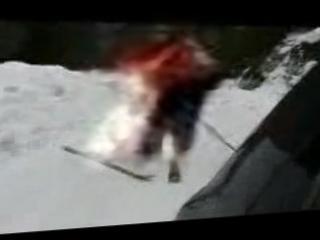} \\
{\footnotesize (c) Future Pose } & {\footnotesize (d) Future Video} \\
\end{tabular}  
\vspace{0.1in}
\caption{In this paper, we train a generative model that takes in (a) an initial clip with (b) a detected pose. Given this information, we generate different motions in (c) pose space using a Variational Autoencoder and utilize a Generative Adversarial Network to generate (d) pixels of the forecast video. Best seen in our \href{http://www.cs.cmu.edu/~jcwalker/POS/POS.html}{videos}.}
\label{teaser}
\vspace{-0.2in}
\end{figure}

Consider the image in Figure~\ref{teaser}. Given the context of the scene and perhaps a few past frames of the video, we can infer what likely action this human will perform. This man is outside in the snow with skis. What is he going to do in the near future? We can infer he will move his body forward towards the viewer. Visual forecasting is a fundamental part of computer vision with applications ranging from human computer interaction to anomaly detection. If computers can anticipate events before they occur, they can better interact in a real-time environment. Forecasting may also serve as a pretext task for representation learning~\cite{DoerschThesis16, Vondrick16, Walker16}.

\begin{figure*}
\centering
\includegraphics[trim={0.0in 1.5in 0.0in 1.0in},clip,width=6.5in, height=2.5in]{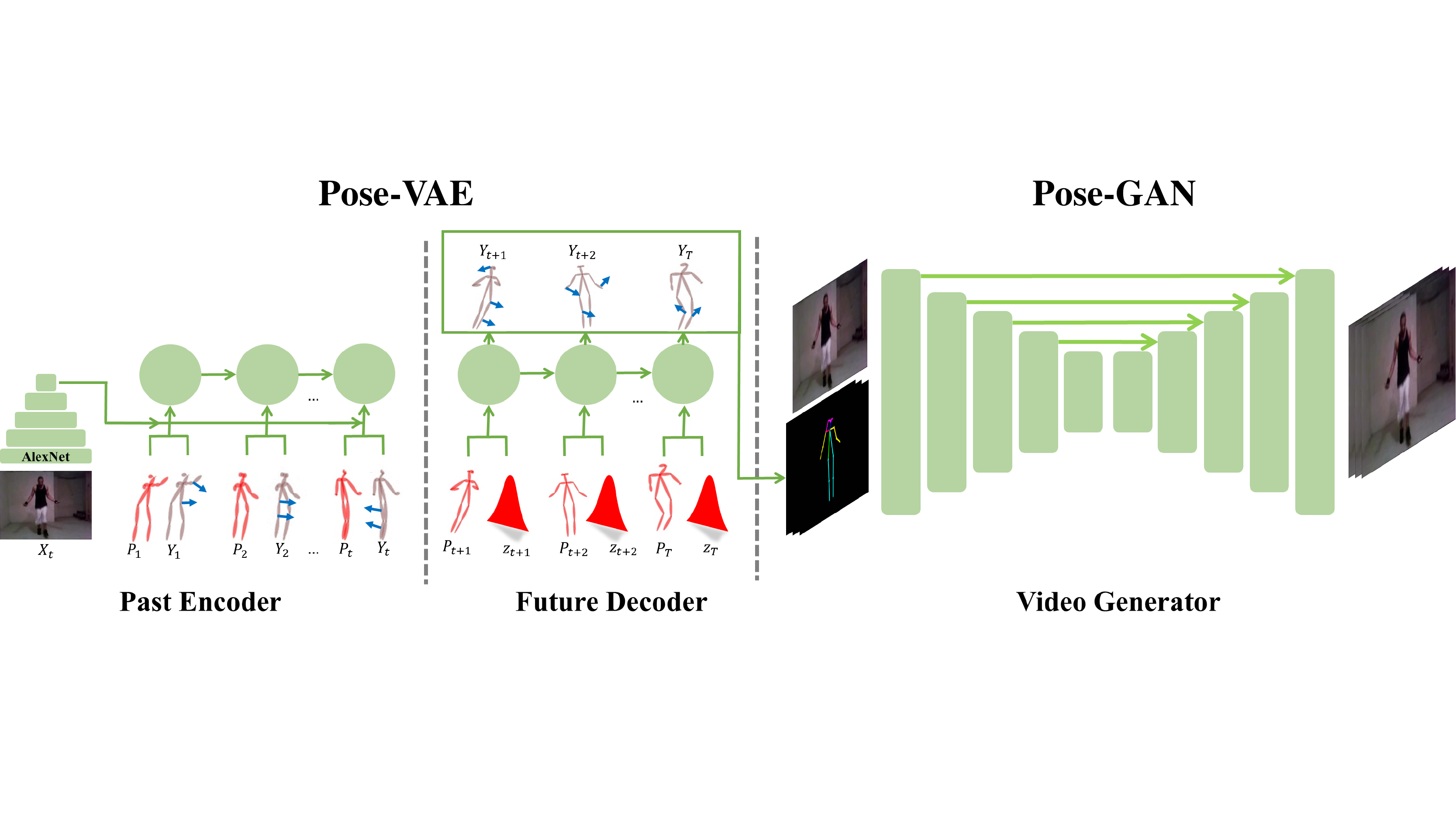}
\vspace{-0.2in}
\caption{Overview of our approach. We use an LSTM, the Past Encoder, to encode the past input into a hidden state. We then input this hidden state into an LSTM with a Variational Autoencoder, the Future Decoder, which predicts future pose velocities based on random samples from latent variables. Given a rendered video of a pose video, we feed this with the input clip into an adversarially trained generator to output the final future video.}
\vspace{-0.2in}
\label{fig:Overview}
\end{figure*}

Given this goal of forecasting, how do we proceed? How can we predict events in a data-driven way without relying on explicit semantic classes or human-labeled data? In order to forecast, we first must determine what is active in the scene. Second, we then need to understand how the structure of the active object will deform and move over time. Finally, 
we need to understand how the pixels will change given the action of the object. All of these steps have a level of uncertainty; however, the second step may have far more uncertainty than the other two. In Figure~\ref{teaser}, we can already tell what is active in this scene, the skier, and given a description of the man's motion, we can give a good guess as to how that motion will play out at the pixel level. He is wearing dark pants and a red coat, so we would expect the colors of his figure to still be fairly coherent throughout the motion. However, the way he skis forward is fairly uncertain. He is moving towards the viewer, but he might move to the left or right as he proceeds. Models that either try to directly forecast pixels ~\cite{Srivastava15, Ranzato14, Xue16, Vondrick16, Matthieu16} or pixel motion~\cite{Walker15, Walker16, Finn16, Pintea14, Yuen10} are forced to perform all of these tasks simultaneously. What makes the problem harder for a complete end-to-end approach is that it has to simultaneously learn the underlying structure (what pixels move together), the underlying physics and dynamics (how the pixels move) and the underlying low-level rendering factors (such as illumination). Forecasting models may instead benefit if they explicitly separate the structure of objects from their low-level pixel appearance.  

The most common agent in videos is a human. In terms of obtaining the underlying structure, there have been major advances in human pose estimation~\cite{Cao17, Wei16, Chu17, Newell16} in images, making 2D human pose a viable ``free'' signal in video. In this paper, we exploit these advances to self-label video and aid forecasting. We propose a new approach to video forecasting by leveraging a more tractable space---human pose---as intermediate representation. Finally, we combine the strengths of VAE with those of GANs. The VAE estimates the probability distribution over future poses given a few initial frames. We can then forecast different plausible events in pose space. Given this structure, we then can use a Generative Adversarial Network to fill in the details and map to pixels, generating a full video. Our approach does not rely on any explicit class labels, human labeling, or any prior semantic information beyond the presence of humans. We provide experimental results that show our model is able to account for the uncertainty in forecasting and generate plausible videos.

\section{Related Work}
\noindent{\bf Activity Forecasting:} Much work in activity forecasting has focused on predicting future semantic action classes~\cite{Savarese14, Ma16, Hoai14} or more generally semantic information~\cite{Vondrick15, RhinehartK16a}. One way to move beyond semantic classes is to forecast an underlying aspect of human activity---human motion. However, the focus in recent work has been in specific data domains such as pedestrian trajectories in outdoor scenes~\cite{Kitani12, Savarese16b, Savarese16c} or pose prediction on human-labeled mocap data~\cite{Jain16,Fragikiadaki15}. In our paper, we aim to rely on as few semantic assumptions as possible and move towards approaches that can utilize large amounts of unlabeled data in unconstrained settings. The only assumption we make on our data is that there is at least one detectable human in the scene. While the world of video consists of more than humans, we find that the great majority of video data in computer vision research focuses on human actions ~\cite{Soomro12, Kuehne11, ActivityNet, Sports1m, THUMOS, Sigurdsson16}.

\noindent{\bf Generative Models:} Our paper incorporates ideas from recent work in generative models of images.
This body of work views images as samples from a distribution and seeks to build parametric models (usually CNNs) that can sample from these distributions to generate novel images.
Variational Autoencoders (VAEs) are one such approach which have been employed in a variety of visual domains. These include modeling faces~\cite{Kingma14a,Rezende14} and handwritten digits~\cite{Kingma14a,Salimans15}. Furthermore, Generative Adversarial Networks (GANs)~\cite{Goodfellow14, Radford15, Isola16, Denton15, Pathak16}, have shown promise as well, generating almost photo-realistic images for particular datasets. There is also a third line of work including PixelCNNs and PixelRNNs~\cite{Oord16a, Oord16b} which model the conditional distribution of pixels given spatial context.
In our paper, we combine the advantages of VAEs with GANs. VAEs are inherently designed to estimate probability distributions of inputs, but utilizing them for estimating pixel distributions often leads to blurry results. On the other hand, GANs can produce sharp results, especially when given additional structure~\cite{Wang16, Isola16, Pathak16, Reed16}. Our VAE estimates a probability distribution over the more tractable space of pose while a GAN conditions on this structure to produce pixel videos.

\noindent{\bf Forecasting Video:} In the last few years there have been a great number of papers focusing specifically on data-driven forecasting in videos. One line of work directly predicts pixels, often incorporating ideas from generative models. Many of these papers used LSTMs~\cite{Srivastava15, Ranzato14, Patraucean16}, VAEs~\cite{Xue16}, or even a PixelCNN approach~\cite{Kalchbrenner16}. While these approaches work well in constrained domains such as moving MNIST characters, they lead to blurring when applied to more realistic datasets. A more promising direction for direct pixel prediction may be the use of adversarial loss~\cite{Vondrick16, Matthieu16}. These methods seem to yield better results for unconstrained, realistic videos, but they still struggle with blurriness and uninterpretable outputs.

Given the difficulty of modeling direct pixels in video, many~\cite{Yuen10, Pintea14, Walker15, Walker16, Brabandere16} have resorted to pixel motion for forecasting. This seems reasonable, as motion trajectories are much more tractable than direct pixel appearances. These approaches can generate interpretable results for short time spans, but over longer time spans they are untenable. They depend on warping existing pixels in the scene. However, this general approach is a conceptual dead-end for video prediction---all it can do is move existing pixels. These methods cannot model occluded pixels coming into frame or model changes in pixel appearance. 

Modeling low-level pixel space is difficult, and motion-based approaches are inherently limited. How then can we move forward with data-driven forecasting? Perhaps we can use some kind of intermediate representation that is more tractable than pixels. One paper~\cite{Walker14} explored this idea using HOG patches as an intermediate representation for forecasting. However, this work focused on specific domains involving cars or pedestrians and could only model rigid objects and rough appearances. In this paper, we use an intermediate representation which is now easy to acquire from video---human pose. Human pose is still visually meaningful, representing interpretable structure for the actions human perform in the visual world. It is also fairly low dimensional---many 2D human pose models only have 18 joints. Estimating a probability distribution over this space is going to be far more tractable than pixels. Yet human pose can still serve as a proxy for pixels. Given a video of a moving skeleton, it is then an easier task to fill in the details and output a final pixel video. We find that training a Video-GAN~\cite{Vondrick16} on completely unconstrained videos leads to results that are many times visually uninterpretable. However, when given prior structure of pose, performance improves dramatically. 
\section{Methodology}
\begin{figure}
\centering
\includegraphics[trim={0.2in 0in 6.5in 0in},clip,width=3.0in, height=3.5in]{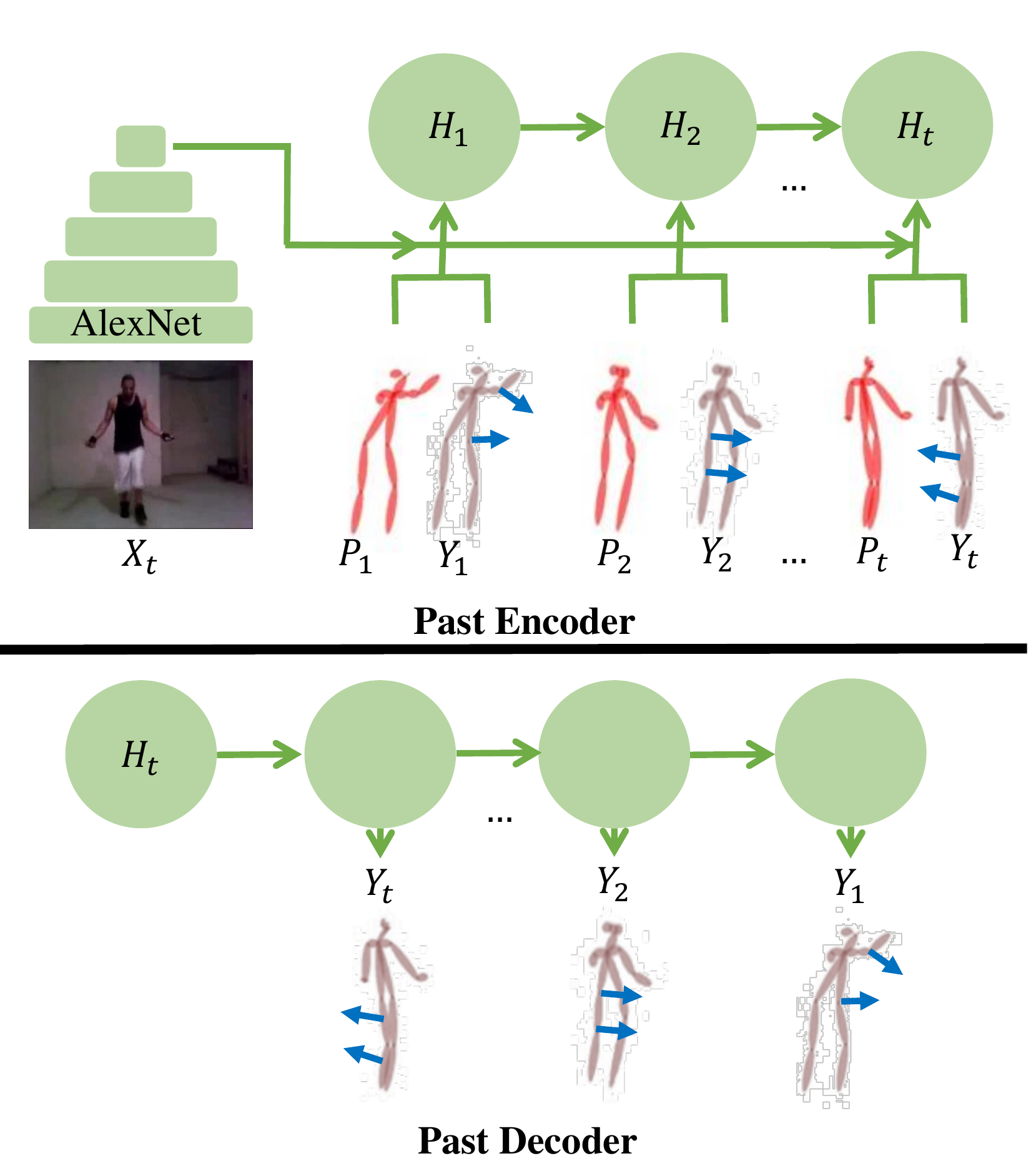} 
\caption{{\bf Past Encoder-Decoder Network.} This portion of Pose-VAE encodes the past input deterministically. The Past Encoder reads in image features from $X_{t}$, corresponding past poses $P_{1..t}$, and their corresponding velocities $Y_{1...t}$. The Past Decoder replays the pose velocities in reverse order. The Past Decoder is only used for training and is discarded during testing. }
\label{fig:SequentialPast}
\end{figure}

In this paper we break down the process of video forecasting into two steps. We first predict the high-level movement in pose space using the \textbf{Pose-VAE}.
Then we use this structure to predict a final pixel level video with the \textbf{Pose-GAN}. 

\subsection{Pose-VAE}
The first step in our pipeline is forecasting in pure pose space. At time $t$, given a series of past poses $P_{1..t}$ and the last frame of in input video $X_{t}$, we want to predict the future poses up to time step $T$, $P_{t+1..T}$. $P_t \in \mathcal{R}^{36}$ is a 2D pose as timestep $t$ represented by the $(x,y)$ locations of $18$ key-points. We actually predict a series of pose velocities $Y_{t+1..T}$.
Given the pose velocities and an initial pose we can then construct the future pose sequence.



To accomplish this forecasting task, we build upon ideas related to sequential encoder-decoder networks ~\cite{Srivastava15, Fragikiadaki15}. As in these papers we can use an LSTM to encode the past information sequence. We call this the \textbf{Past Encoder} which takes in the past information $X_t$, $P_{1..t}$, and $Y_{1..t}$ and encodes it in a hidden representation $H_t$. We also have \textbf{Past Decoder} module to reconstruct the past information from the hidden state. Given this encoding $H_t$ of the past, it would be tempting to use another LSTM to simply produce the future sequence of poses similar to~\cite{Srivastava15}. However, forecasting the future is not a deterministic problem; there may be multiple plausible outcomes of a video. Forecasting actually requires estimating a probability distribution over possible events. To solve this problem, we use a probabilistic Future Decoder. Our probabilistic decoder is nothing but a conditional variational autoencoder where the future velocity $Y_{t+1}$ is predicted given the past information $H_t$, the current pose $P_{t+1}$ (estimated from $P_t$ and $Y_t$), and the random latent vector $z_{t+1}$. The hidden states of the Future Decoder are updated using the standard LSTM update rules.

\noindent{\bf Variational Autoencoders:} A Variational Autoencoder~\cite{Kingma14a} attempts to estimate the probability distribution $P(Y|z)$ of its input data $Y$ given latent variables $z$. An encoder $Q(z|Y)$  learns to encode the inputs into a stochastic latent variable $z$. The decoder $P(Y|z)$ then reconstructs the inputs based on what is sampled from $z$. During training, $z$ is regularized to match $\mathcal{N}(0,1)$ through KL-Divergence. During testing we can then sample our distribution of $Y$ by first sampling $z \sim \mathcal{N}(0,1)$ and then feeding our sample through a neural network $P(Y|z)$ to create a sample from the distribution of Y. Another interpretation is that the decoder $P$ transforms the latent random variable $z \sim \mathcal{N}(0,1)$ into random variable $Y \sim P(Y|z)$.

In our case, we want to estimate a distribution of future pose velocities given the past. Thus we aim to ``encode'' the future into latent variables $z = [z_{t+1}, z_{t+2}, ... z_T]$. Concretely, we wish to learn a way to estimate the distribution $P(Y_{t+1..T}|z,H_{t})$ of future pose velocities $Y_{t+1..T}$ given our encoded knowledge of the past $H_{t}$. Thus we need to train a ``Future Encoder'' that learns an encoding for latent variables $z \sim Q(z|Y_{t+1..T},H_{t})$, where $Q$ is trained to match $\mathcal{N}(0,1)$ as closely as possible. 
During testing, as in~\cite{Walker16}, we sample $z \sim \mathcal{N}(0,1)$ and feed sampled $z$ values into the future decoder network to output different possible forecasts.

\begin{figure}
\centering
\includegraphics[trim={0.5in 0in 6.5in 0in},clip,width=3.0in, height=3.5in]{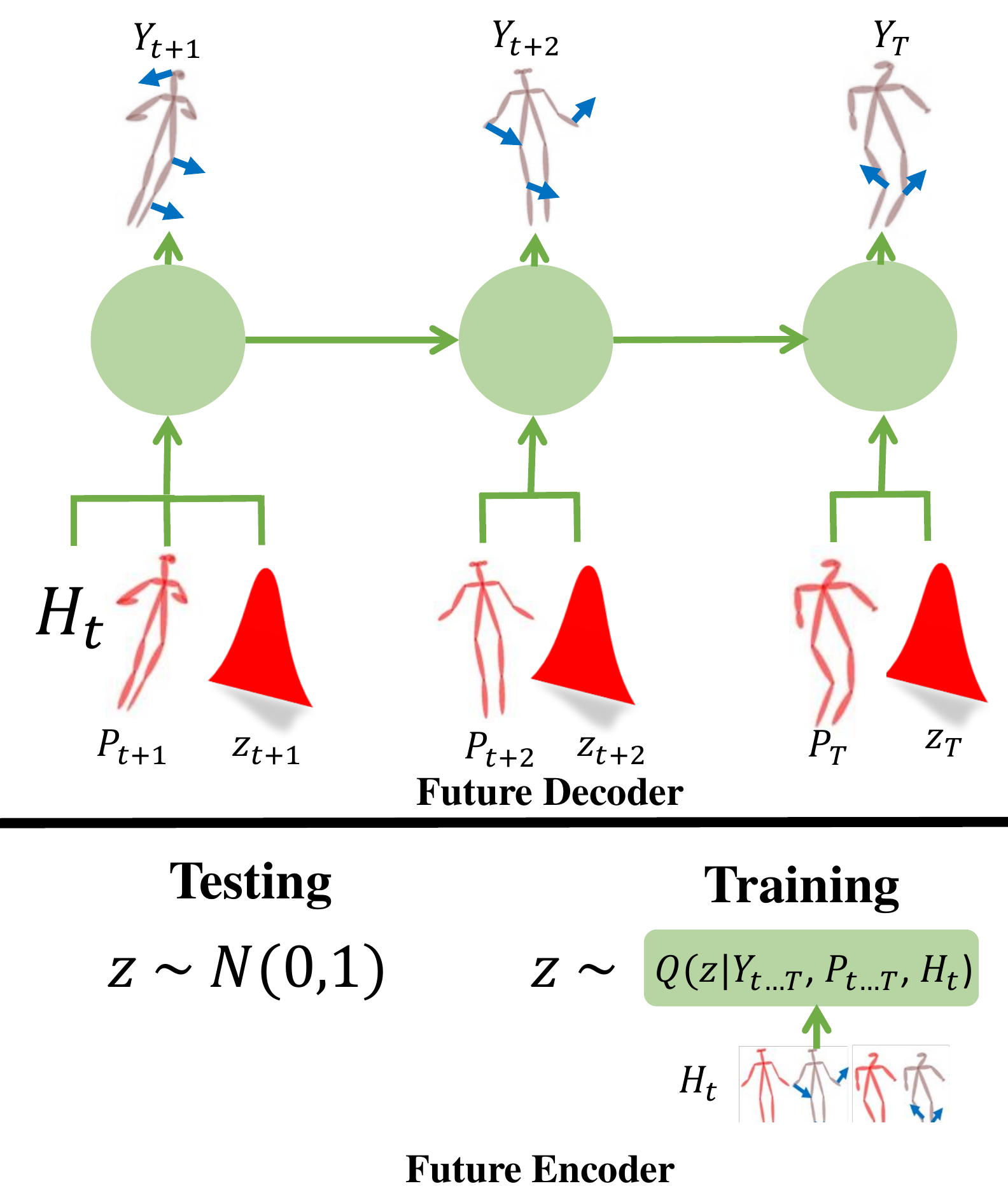} 
\caption{{\bf Future Encoder-Decoder Network.} This portion of Pose-VAE encodes the future stochastically. The Future Encoder is a Variational Autoencoder which takes the past $H_{t}$ and the future pose information $Y_{t+1...T}, P_{t+1...T}$ as input and outputs a Normal Distribution $Q$. The Future Decoder then samples $z$ from $Q$ to reconstruct the pose motions $Y_{t+1...T}$ given past $H_{t}$ and poses $P_{t+1....T}$. During testing, the future is not known, so the Future Encoder is discarded, and only the Future Decoder is used with $z \sim \mathcal{N}(0,1)$. }
\label{fig:SequentialFuture}
\end{figure}

\noindent{\bf Past Encoder-Decoder:} Figure~\ref{fig:SequentialPast} shows the Past Encoder-Decoder. The Past Encoder takes as input a frame $X_t$, a series of previous poses $P_{1..t}$, and the previous pose velocities $Y_{1..t}$. We apply a convolutional neural network on $X_t$. The units from the pose information and the image features are concatenated and then fed into an LSTM. After encoding the entire sequence, we use the hidden state of the LSTM at step $t$, $H_t$ to condition the Future Decoder. To enforce that $H_t$ encodes the pose velocity, the hidden state of of the encoding LSTM is fed into a decoder LSTM, the Past Decoder, which is trained through Euclidean loss to reconstruct $Y_{1..t}$ in reverse order. This enforces that the network learns a ``memory'' of past inputs~\cite{Srivastava15}. The Past Decoder exists only as an aid for training, and at test time, only the Past Encoder is used.

\noindent{\bf Future Encoder-Decoder:} Figure~\ref{fig:SequentialFuture} shows the Future Encoder-Decoder. The Future Encoder-Decoder is composed of a VAE encoder (Future Encoder) and a VAE decoder (Future Decoder) both conditioned on past information $H_{t}$. The Future Encoder takes the future pose velocity $Y_{t+1..T}$ and the past information $H_t$ and encodes it as a mean and variance $\mu(Y_{t+1..T}, H_t)$ and $\sigma(Y_{t+1..T}, H_t)$. We then sample a latent variable $z \sim Q(z|Y_{t+1..T}, H_t) = \mathcal{N}(\mu,\sigma)$. During testing, we 
sample $z$ from a standard normal, so during training we incorporate a KL-divergence loss such that $Q$ matches $\mathcal{N}(0,1)$ as closely as possible. Given the latent variable $z$ and the past information $H_t$, the Future Decoder recovers an approximation of the future pose sequence $\hat{Y}_{t+1..T}(z, H_t)$. The training loss for this network is the usual VAE Loss. It is Euclidean distance from the pose trajectories combined with KL-divergence loss of $Q$ from $\mathcal{N}(0,1)$.

\vspace{-.3cm}
\begin{equation}
\begin{split}
L(\hat{Y}_{t+1..T},Y_{t+1..T}) = ||Y_{t+1..T} - \hat{Y}_{t+1..T}||^{2} +
\\
\lambda\mathcal{KL}\left[Q(z|Y_{t+1..T},H_t)\|\mathcal{N}(0,1)\right]
\label{eq:VAE}
\end{split}
\end{equation}
\vspace{-.3cm}

At every future step $t_f$, the Future Decoder takes in $z_{t_f}$, as well as the current pose $P_{t_f}$ and outputs the pose motion $Y_{t_f}$. At training time, we use the ground truth poses, but at test time, we recover the future poses by simply adding the pose trajectory information $P_{t_f+1}=P_{t_f}+Y_{t_f}$. 

\noindent{\bf Implementation Details:} We train our network with Adam Solver at a learning rate of 0.001 and $\beta_{1}$ of 0.9. For the KL-divergence loss we set $\lambda=0.00025$ for 60000 iterations and then set $\lambda=0.0005$ for an additional 20000 iterations of training. Every timestep $t$ represents 0.2 second. We conditioned the past on 2 timesteps and predict for 5 timesteps. For the convolutional network over the image network, we used an architecture almost identical to AlexNet~\cite{Krizhevsky12} with the exception of a smaller (7x7) receptive field at the bottom layer and the addition of batch normalization layers. All layers in the entire network were trained from scratch. The LSTM units consist of two layers, both 1024 units. The Future Encoder is a simple single hidden layer network with ReLU activations and a hidden size of $512$. 


\subsection{Pose-GAN}

\noindent{\bf Generative Adversarial Networks:} Once we sample a pose prediction from our Pose-VAE, we can then render a video of a moving skeleton. Given an input image and a
video of the skeleton, we train a Generative Adversarial Network to predict a pixel level video of future events. As described in ~\cite{Wang16}, GANs consist of two models pitted against each other: a generator $G$ and a discriminator $D$. The generator $G$ takes the input skeleton video and image and attempts to generate a realistic video. The discriminator $D$, trained as a binary classifier, attempts to classify videos as either real or generated. During training, $G$ will try to generate videos which fool $D$, while $D$ will attempt to distinguish the fake videos generated by $G$ from ones sampled from the future video frames.
Following the work of ~\cite{Isola16, Vondrick16} we do not use any noise variables for the adversarial network. All the noise is contained in the Pose-VAE through $z$.

The loss for discriminator $D$ is:

\vspace{-.3cm}
\begin{equation}
\begin{split}
L_{D} = \sum_{i=1}^{M/2} l(D(V_{i}),l_r)+\hspace{-.3cm}
\sum_{i=M/2+1}^{M} l(D(G(I, S_{T})),l_f)
\end{split}
\end{equation}
\vspace{-.3cm}

Where $V$ are videos, $M$ is the batch size, $I$ is an input image, and $S_{T}$ is a video of a pose skeleton, $l_r$ is the real label ($1$), and $l_f$ is the fake label ($0$). Inside the batch $M$, half of videos $V$ are generated, and the rest are real. The loss function $l$ here is the binary entropy loss. 


The loss for generator $G$ is:

\vspace{-.5cm}
\begin{equation}
\begin{split}
L_{G} \hspace{-.1cm} = \hspace{-.4cm} \sum_{i=M/2+1}^{M} \hspace{-.2cm} l(D(G(I, S_{T})),l_r) +
\vspace{-10.0cm} \alpha||G(I, S_{T}) - V_{i}||_1
\end{split}
\end{equation}
\vspace{-.5cm}

\begin{figure}
\centering
\includegraphics[trim={1in 1.2in 1in 1.5in},clip,width=3.25in, height=1.5in]{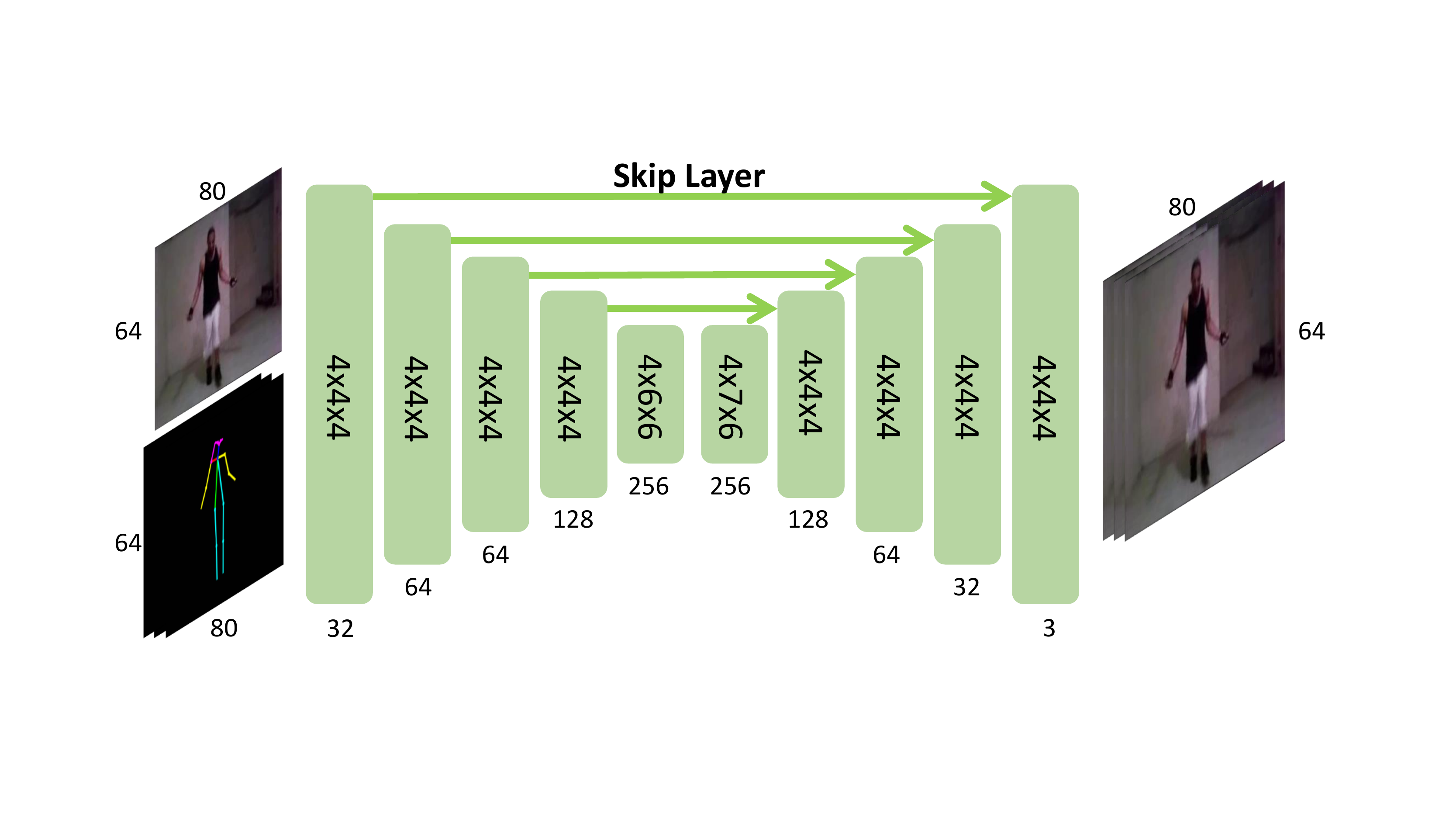} 
\vspace{-0.2in}
\caption{{\bf Generator Architecture}. We use volumetric convolutions at each layer. Receptive field size represents time, width, and length. For each frame in the input pose video we stack the input frame as an extra 3 channels, making each input frame 80x64x6. The number of input and output frames is 32. The output consists of 32 frames, 80x64 pixels.}
\vspace{-0.2in}
\label{fig:GANArchitecture}
\end{figure}

Given our Pose-VAE, we can now generate plausible pose motions given a very short clip input. For each sample, we can render a video of a skeleton visualizing how a human will deform over the last frame of the input image. Recent work in adversarial networks has shown that GANs benefit from given structure~\cite{Isola16, Pathak16, Reed16}. In particular,~\cite{Reed16} showed that GANs improve on generating humans when given initial keypoints. In this paper we build on this work by extending this idea to Conditional Video GANs. Given an image and a generated skeleton video, we train a GAN to generate a realistic video at the pixel level. Figure~\ref{fig:GANArchitecture} shows the Pose-GAN network. The architecture of the discriminator $D$ is nearly identical to that of~\cite{Vondrick16}. 

\noindent{\bf Implementation Details:} The Pose-GAN consists of five volumetric convolutional layers with receptive fields of 4, stride of 2, and padding of 1. At each layer LeakyReLU units and Batch Normalization are used. The only difference is that the input is a 64x80 video. For the generator $G$, we first encode the input using a series of five Volumentric Convolutional Layers with receptive fields of 4, stride of 2, and padding of 1. We use LeakyReLU and Batch Normalization at each layer. In order to handle the modified aspect ratio of the input (80x64), the fifth layer has a receptive field of 6 in the spatial dimensions. The top five layers are the same but in reverse, gradually increasing the spatial and temporal resolution to 64x80 pixels at 32 frames. Our training parameters are identical to ~\cite{Vondrick16}, except that we set our regularization parameter $\alpha=1000$. Similar to~\cite{Isola16}, we utilize skip layers for the top part of the network. For the top five layers, ReLU activation and Batch Normalization is used. The final layer is sent through a TanH function in order to scale the outputs. 

\section{Experiments}
We evaluate our model on UCF-101~\cite{Soomro12} in both pose space and video space. We utilized the training split described in~\cite{Walker15} which uses a large portion for training data. This split leaves out one video group for testing and uses the rest for training. In total we use around 1500 one-second clips for testing. To label the data we utilize the pose detector of Cao et al.~\cite{Cao17} and use the videos above an average confidence threshold. We perform temporal smoothing over the pose detections as a post-processing step.

\subsection{Pose Evaluation}

First we evaluate how well our Pose-VAE is able to forecast actions in pose space. There has been some prior work~\cite{Fragikiadaki15, Jain16} on forecasting pose in mocap datasets such as the H3.6M dataset~\cite{Ionescu14}. However, to the best of our knowledge there has been no evaluation on 2D pose forecasting on unconstrained, realistic video datasets such as UCF101. We compare our Pose-VAE against state-of-the-art baselines. First, we study the effects of removing the VAE from our Future Decoder. In that case, the forecasting model becomes a Encoder-Recurrent-Decoder network similar to~\cite{Fragikiadaki15}. We also implemented a deterministic Structured RNN model~\cite{Jain16} for forecasting with LSTMs explictly modeling arms, legs and torso.
Finally, we take a feed-forward VAE~\cite{Walker16} and apply it to pose trajectory prediction. In our case, the feed-forward VAE is conditioned on the image and past pose information, and it only predicts pose trajectories. 

\noindent{\bf Quantitative Evaluations:} For evaluation of pose forecasting, we utilize Euclidean distance from the ground-truth pose velocities. However, specifically taking the Euclidean distance over all the samples from our model in a given clip may not be very informative. Instead, we follow the evaluation proposed by ~\cite{Walker16}. For a set number of samples $n$, we see what is the best possible prediction made by the model and consider the error of closest sample from the ground-truth. We then measure how this minimum error changes as the sample size $n$ increases and the model is given more chances. We make our deterministic baselines stochastic by treating the output as a mean of a multivariate normal distribution. For these baselines, we derive the bandwidth parameters from the variance of the testing data. Attempting to use the optimal MLE bandwidth via gradient search led to inferior performance. We describe the possible reasons for this phenomenon in the results section.  

\subsection{Video Evaluation}

We also evaluate the final video predictions of our method. These evaluations are far more difficult as pixel space is much higher-dimensional than pose space. However, we nonetheless provide quantitative and qualitative evaluations to compare our work to the current state of the art in pixel video prediction. Specifically, we compare our method to Video-GAN~\cite{Vondrick16}. For this baseline, we only make two small modifications to the original architecture---Instead of a single frame, we condition Video-GAN on 16 prior frames. We also adjust the aspect ratio of the network to output a 64x80 video.

\noindent{\bf Quantitative Evaluations:} To evaluate the videos, we use the Inception score, first introduced in~\cite{Salimans16}. In the original method, the authors use the Inception model~\cite{szegedy2015going} to get a conditional label distribution for their generated images. In our case, we are generating videos, so we use a two-stream action classifier~\cite{LWang15} to get a conditional label distribution $p(y|x)$ where $x$ is our generated video and $y$ is the action class. We calculate the label distribution by taking the average classification output of both the rgb and flow stream in the classifier. As in~\cite{Salimans16}, we use the metric $\exp(\mathbb{E}_x KL(p(y|x)||p(y))$. In our case, our $x$ is generated from an input video sequence $fr$ and in some models a latent variable $z$, giving us the metric $\exp(\mathbb{E}_{fr, z} KL(p(y|x=G(f,z))||p(y))$. The intuition behind the metric is diversity; if a given classifier is highly confident of particular classes in the generated videos, then the Inception score will be large. If it has low confidence and is unsure what classes are in the videos, the conditional distribution will be close to the prior and the Inception score will be low.

We also propose a new evaluation metric based on the test statistic Maximum Mean Discrepancy (MMD)~\cite{Gretton14}. MMD was proposed as a test statistic for a two sample test---given samples drawn from two distributions $\textbf{P}$ and $\textbf{Q}$, we test whether or not the two distributions are equal.  

While the MMD metric is based on a two sample test, and thus is a metric for how similar the generated distribution is from the ground truth, the Inception score is a rather an ad hoc metric measuring the entropy of the conditional label distribution and marginal label distribution. We present scores for both metrics, but we believe MMD to be a more statistically justifiable metric

 \begin{figure}[t!]
\centering
\includegraphics[width=2.75in, height=2.0in]{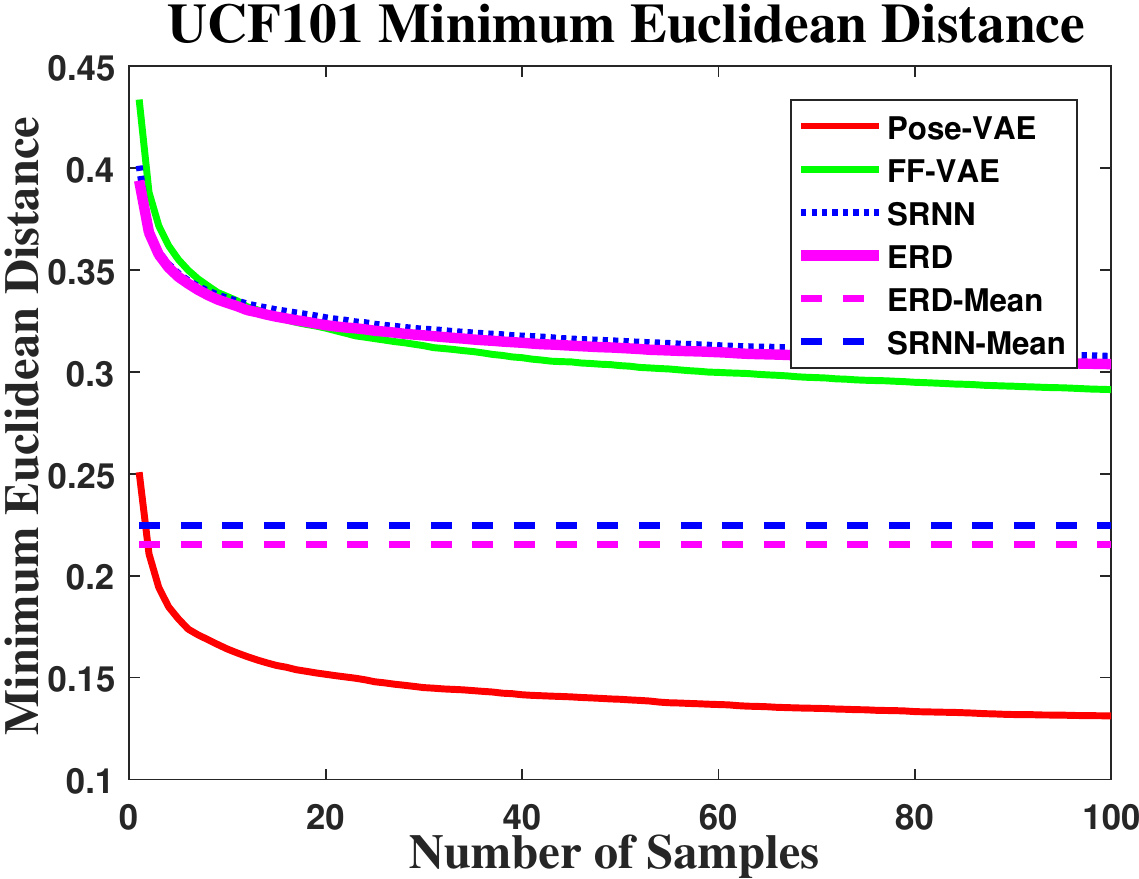} 
\caption{Here we show Minimum Euclidean Distance averaged over the testing examples. We take this nearest prediction in each example and plot the average of the error as the number of samples grows.}
\label{fig:EuclideanErrorUCF101}
\end{figure}

The exact MMD statistic for a class of functions $\mathcal{F}$ is:
\begin{equation}
\label{eqMMD}
MMD[\mathcal{F}, \textbf{P}, \textbf{Q}] = \underset{f \in \mathcal F}{\sup}(\mathbb{E}_{x \thicksim \textbf{P}}[f(x)]-\mathbb{E}_{y \thicksim \textbf{Q}}[f(y)]).
\end{equation}
Two distributions are equal if and only if for all functions $f \in \mathcal F$, $\mathbb{E}_x[f(x)]=\mathbb{E}_y[f(y)]$, so if $\textbf{P} \overset{d}{=} \textbf{Q}$, $MMD = 0$ where $\mathcal{F}$ is the set of all functions. Since evaluating over the set of all functions is intractable, we instead evaluate for all functions in a Reproducing Kernel Hilbert Space to approximate. We use the unbiased estimator for MMD from~\cite{Gretton14}.

Some nice properties of this test statistic are that the empirical estimate is consistent and converges in $O(\frac{1}{\sqrt n})$ where $n$ is the sample size. This is independent of the dimension of data \cite{Gretton14}. MMD has been used in generative models before, but as part of the training procedure rather than as an evaluation criteria. \cite{Dziugaite15} uses MMD as a loss function to train a generative network to produce better images. \cite{Li15} extends this by first training an autoencoder and then training the generative network to minimize MMD in the latent space of the autoencoder, achieving less noisy images.

We choose Gaussian kernels with bandwidth ranging from $10^{-4}$ to $10^9$ and choose the maximum of the values generated from these bandwidths as the reported value since from eq. (\ref{eqMMD}), we want the maximum distance out of all possible functions.

 Like Inception score, we use semantic features instead of raw pixels or flow for comparison. However, we use the $fc7$ feature space rather than the labels. We concatenate the $fc7$ features from the rgb stream and the flow stream of our action classifier.  This choice choice of semantic $fc7$ features is supported by the results in \cite{Li15} which show that training MMD on a lower-dimensional latent space rather than the original image space generates better looking images.
 
\begin{table}
\parbox{.45\linewidth}{
\centering
\captionsetup{justification=centering}
\caption{Inception Scores. Higher is better.}
    \begin{tabular}{|c|c|}
         \hline 
         Method & Inception\\ \hline
         Real & 3.81$\pm$ 0.04 \\ \hline
         Ours & 3.14$\pm$ 0.04\\ \hline
        ~\cite{Vondrick16} & 1.74$\pm$ 0.01\\ \hline
    \end{tabular}
    \label{Inception}
}
\hfill
\parbox{.45\linewidth}{
\centering
\captionsetup{justification=centering}
\caption{MMD Scores. Lower is better.}
    \begin{tabular}{|c|c|}
         \hline 
         Method & MMD\\ \hline 
         Real & 0.003$\pm$ 0.0003\\ \hline
        Ours & 0.022$\pm$ 0.0001\\ \hline
        ~\cite{Vondrick16} & 0.139$\pm$ 0.0001 \\ \hline
     \end{tabular}
     \label{MMD}
}
\end{table}

\begin{figure*}[t!]
\centering
\begin{tabular}{ ccccc }
\includegraphics[height=0.12\textwidth,width=0.16\textwidth]{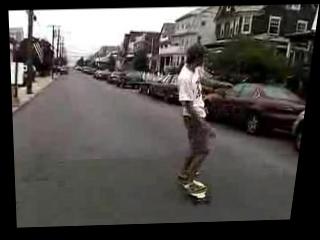} &
\includegraphics[height=0.12\textwidth,width=0.16\textwidth]{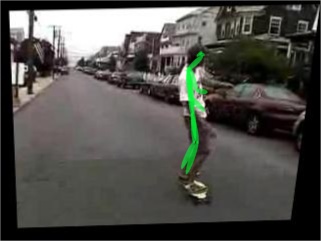} &
\includegraphics[height=0.12\textwidth,width=0.16\textwidth]{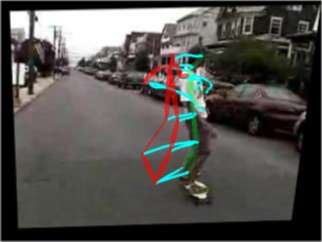} &
\includegraphics[height=0.12\textwidth,width=0.16\textwidth]{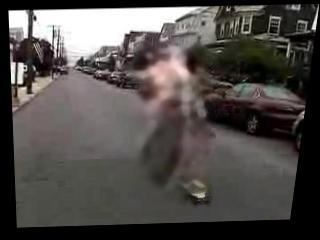} &
\includegraphics[height=0.12\textwidth,width=0.16\textwidth]{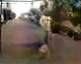} \\
\includegraphics[height=0.12\textwidth,width=0.16\textwidth]{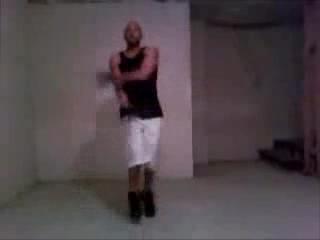} &
\includegraphics[height=0.12\textwidth,width=0.16\textwidth]{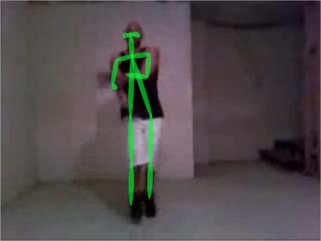} &
\includegraphics[height=0.12\textwidth,width=0.16\textwidth]{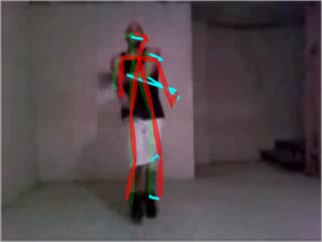} &
\includegraphics[height=0.12\textwidth,width=0.16\textwidth]{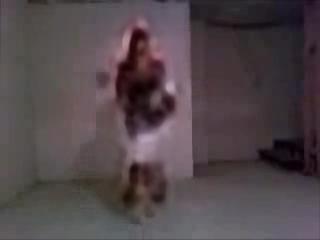} &
\includegraphics[height=0.12\textwidth,width=0.16\textwidth]{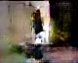}
\\
\includegraphics[height=0.12\textwidth,width=0.16\textwidth]{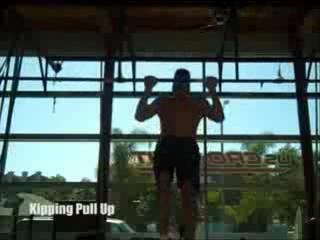} &
\includegraphics[height=0.12\textwidth,width=0.16\textwidth]{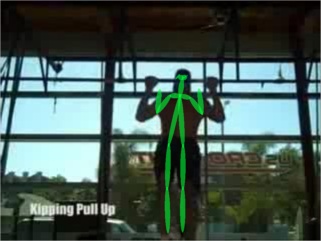} &
\includegraphics[height=0.12\textwidth,width=0.16\textwidth]{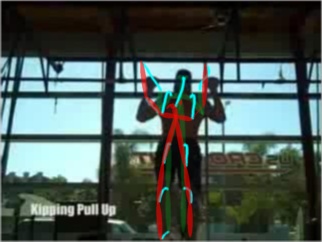} &
\includegraphics[height=0.12\textwidth,width=0.16\textwidth]{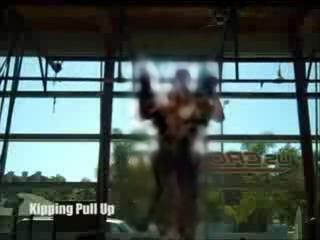} &
\includegraphics[height=0.12\textwidth,width=0.16\textwidth]{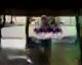}
\\
\includegraphics[height=0.12\textwidth,width=0.16\textwidth]{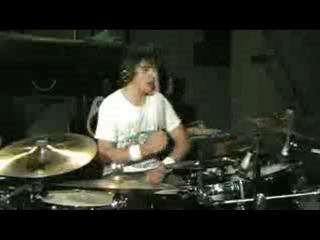} &
\includegraphics[height=0.12\textwidth,width=0.16\textwidth]{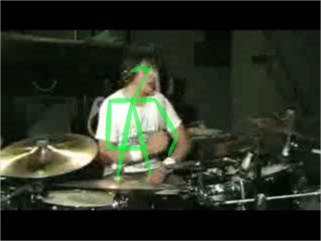} &
\includegraphics[height=0.12\textwidth,width=0.16\textwidth]{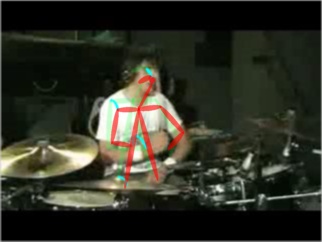} &
\includegraphics[height=0.12\textwidth,width=0.16\textwidth]{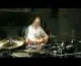} &
\includegraphics[height=0.12\textwidth,width=0.16\textwidth]{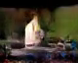}
\\
\includegraphics[height=0.12\textwidth,width=0.16\textwidth]{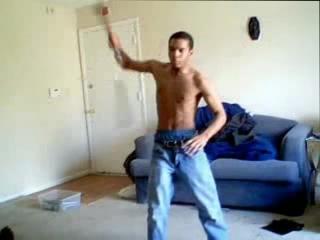} &
\includegraphics[height=0.12\textwidth,width=0.16\textwidth]{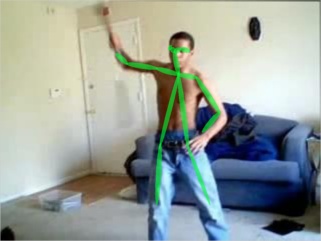} &
\includegraphics[height=0.12\textwidth,width=0.16\textwidth]{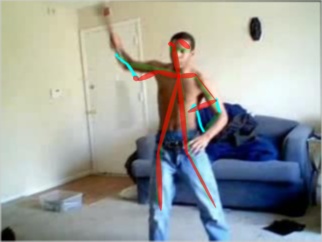} &
\includegraphics[height=0.12\textwidth,width=0.16\textwidth]{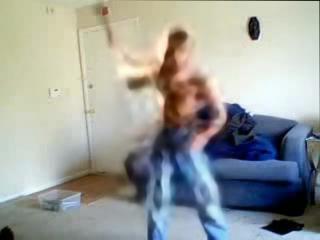} &
\includegraphics[height=0.12\textwidth,width=0.16\textwidth]{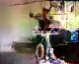}
\\
\includegraphics[height=0.12\textwidth,width=0.16\textwidth]{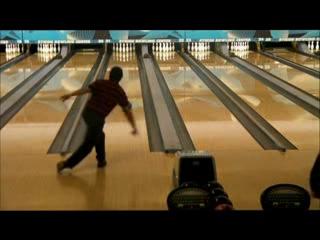} &
\includegraphics[height=0.12\textwidth,width=0.16\textwidth]{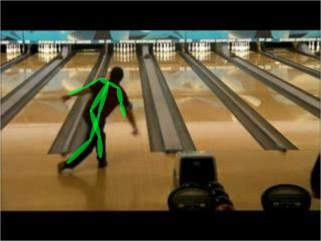} &
\includegraphics[height=0.12\textwidth,width=0.16\textwidth]{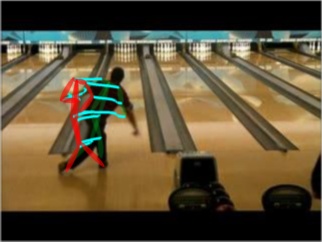} &
\includegraphics[height=0.12\textwidth,width=0.16\textwidth]{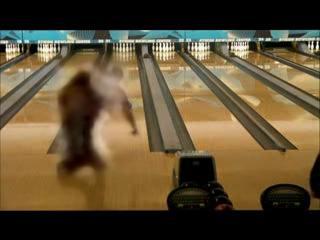} &
\includegraphics[height=0.12\textwidth,width=0.16\textwidth]{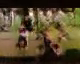}
\\
{(a) Input Clip } & {(b) Input Pose} & {(c) Future Pose} & {(d) Our Forecast} & {(e)~\cite{Vondrick16} Forecast}\\
\end{tabular}  
\vspace{0.1in}
\caption{Here are some selected qualitative results from our model. Given an input clip (a) and a set of poses (b), we forecast a future pose motion (c) and then use this structure to predict video (d). These pose motions represent the largest cluster of samples from Pose-VAE for each input. Best seen in our \href{http://www.cs.cmu.edu/~jcwalker/POS/POS.html}{videos}.}
\label{fig:qualitative}
\vspace{-0.2in}
\end{figure*}


\section{Results}
\subsection{Qualitative Results}
In Figure~\ref{fig:qualitative} we show the qualitative results of our model. The results are best viewed as videos; we strongly encourage readers to look at our \href{http://www.cs.cmu.edu/~jcwalker/POS/POS.html}{videos}. In order to generate these results, for each scene we took 1000 samples from Pose-VAE and clustered the samples above a threshold into five clusters. The pose movement shown is the largest discovered cluster. We then feed the last input frame and the future pose movement into Pose-GAN to generate the final video. On the far right we show the last predicted frame by Pose-GAN. We find that our Pose-GAN is able to forecast a plausible motion given the scene. The skateboarder moves forward, and the man in the second row, who is jumproping, moves his arms to the right side. The man doing a pullup in the third row moves his body down. The drummer plays the drums, the man in the living room moves his arm down, and the bowler recovers to standing position from his throw. We find that our Pose-GAN is able to extrapolate the pixels based on previous information. As the body deforms, the general shading and color of the person is preserved in the forecasts. We also find that Pose-GAN, to a limited extent, is able to inpaint occluded background as humans move from their starting position. In Figure~\ref{fig:qualitative} we show a side-by-side qualitative comparison of our video generation to conditional Video-GAN. While Video-GAN shows compelling results when specifically trained and tested on a specific scene category~\cite{Vondrick16}, we discover that this approach struggles to generate interpretable results when trained on inter-class, unconstrained videos from the UCF101. We specifically find that~\cite{Vondrick16} fails to capture even the general structure of the original input clip in many cases. 

\subsection{Quantitative Results}

We show the results of our quantitative evaluation on pose prediction in Figure~\ref{fig:EuclideanErrorUCF101}. We find our method is able to outperform the baselines on Euclidean distance even with a small number of samples. The dashed lines for ERD and SRNN use the only the direct output as a mean---identical to sampling with variance 0. As expected, the Pose-VAE has a higher error with only a few samples, but as samples grow the error quickly decreases due to the stochastic nature of future pose motion. The solid lines for ERD and SRNN treat the output as a mean of a multivariate normal with variance derived from the testing data. Using the variance seems to worsen performance for these two baselines. This suggests that these deterministic baselines output one particularly incorrect motion for each of the examples, and the distribution of pose motion is not well modeled by Gaussian noise.  We also find our recurrent Pose-VAE outperforms Feedforward-VAE~\cite{Walker16}. Interestingly, FF-VAE underperforms the mean of the two deterministic baselines. This is likely due to the fact that FF-VAE is forced to predict all timesteps simultaneously, while recurrent models are able to predict more refined motions in a sequential manner. 

In Table~\ref{Inception} we show our quantitative results of pixel-level video prediction against~\cite{Vondrick16}. As the Inception score increases, the KL-Divergence between the prior distribution of labels and the conditional class label distribution given generated videos increases. Here we are effectively measuring how often the two stream action classifier detects particular classes with high confidence in the generated videos. We compute variances using bootstrapping. We find, not surprisingly, that real videos show the highest Inception score. In addition, we find that videos generated by our model have a higher Inception score than~\cite{Vondrick16}. This suggests that our model is able to generate videos which are more likely to have particular meaningful features detected by the classifier. In addition to Inception scores, we show the results of our MMD metric in  Table~\ref{MMD}. While Inception is measuring diversity, MMD is instead testing something slightly different. Given the distribution of two sets, we perform a statistical test measuring the difference of the distributions. We again compute a variance with bootstrapping.  We find that, compared to the distribution of real videos, the distribution videos generated by~\cite{Vondrick16} are much further than the videos generated by ours.

\vspace{-0.08in}
\section{Conclusion and Future Work}
In this paper, we make great steps in pixel-level video prediction by exploiting pose as an essentially free source of supervision and combining the advantages of VAEs, GANs and recurrent networks. Rather than try to model the entire scene at once, we predict the high level dynamics of the scene by predicting the pose movements of the humans in the scenes with a VAE and then predict each pixel with a GAN.
We find that our method is able to generate a distribution of plausible futures and outperform contemporary baselines. There are many future directions from this work. One possibility is to combine VAEs with the power of structured RNNs to improve performance. Another direction is to apply our model to representation learning for action recognition and early action detection; our method is unsupervised and thus could scale to large amounts of unlabeled video data. 

\noindent{\bf Acknowledgements:}  We thank the NVIDIA Corporation for the donation of GPUs for this research.  In addition, this work was supported by NSF grant IIS1227495.

{\small
\bibliographystyle{ieee}
\bibliography{egbib}
}

\end{document}